\documentclass[preprint,12pt]{elsarticle}

\usepackage{amssymb}
\usepackage{amsmath}
\usepackage{times}
\usepackage{soul}
\usepackage{url}
\usepackage[hidelinks]{hyperref}
\usepackage[utf8]{inputenc}
\usepackage[small]{caption}
\usepackage{graphicx}
\usepackage{amsmath}
\usepackage{amsthm}
\usepackage{booktabs}
\usepackage{algorithm}
\usepackage{algorithmic}
\usepackage[switch]{lineno}
\usepackage{subfiles}
\usepackage{multirow}
\usepackage{tabularx}
\usepackage{amssymb}
\usepackage{color}
\usepackage{dutchcal}
\usepackage{float}

\begin{document}

\begin{frontmatter}



\title{Zero-Shot Learning with Subsequence Reordering Pretraining for Compound-Protein Interaction}


\author[label1]{Hongzhi Zhang} 
\author[label1]{Zhonglie Liu} 
\author[label1]{Kun Meng} 
\author[label1]{Jiameng Chen} 
\author[label3]{Jia	Wu}
\author[label1]{Bo Du}
\author[label2]{Di Lin\corref{mycorrespondingauthor}}
\author[label2]{Yan Che\corref{mycorrespondingauthor}}
\author[label1,label4]{Wenbin Hu\corref{mycorrespondingauthor}} 
\cortext[mycorrespondingauthor]{Corresponding author}
\affiliation[label1]{organization={School of Computer Science, Wuhan University},
            city={Wuhan},
            country={China}}
\affiliation[label2]{organization={Engineering Research Center for Big Data Application in Private Health Medicine of Fujian Universities, Putian University},
            city={Putian},
            country={China}}
\affiliation[label3]{organization={Macquarie University},
            city={Sydney},
            country={Australia}}
\affiliation[label4]{organization={Wuhan University Shenzhen Research Institute},
            city={Shenzhen},
            country={China}}
\begin{abstract}
    Given the vastness of chemical space and the ongoing emergence of previously uncharacterized proteins, zero-shot compound-protein interaction (CPI) prediction better reflects the practical challenges and requirements of real-world drug development.
    Although existing methods perform adequately during certain CPI tasks, they still face the following challenges: 
    (1) Representation learning from local or complete protein sequences often overlooks the complex interdependencies between subsequences, which are essential for predicting spatial structures and binding properties. 
    (2) Dependence on large-scale or scarce multimodal protein datasets demands significant training data and computational resources, limiting scalability and efficiency.
    To address these challenges, we propose a novel approach that pretrains protein representations for CPI prediction tasks using subsequence reordering, explicitly capturing the dependencies between protein subsequences. Furthermore, we apply length-variable protein augmentation to ensure excellent pretraining performance on small training datasets.    
    To evaluate the model's effectiveness and zero-shot learning ability, we combine it with various baseline methods. The results demonstrate that our approach can improve the baseline model's performance on the CPI task, especially in the challenging zero-shot scenario.
    Compared to existing pre-training models, our model demonstrates superior performance, particularly in data-scarce scenarios where training samples are limited.
    Our implementation is available at \href{https://github.com/Hoch-Zhang/PSRP-CPI}{https://github.com/Hoch-Zhang/DrugDiscovery-DTI/}.
\end{abstract}




\end{frontmatter}



\section{Introduction}

\begin{figure}[h!]
  \centering
  \includegraphics[width=\linewidth]{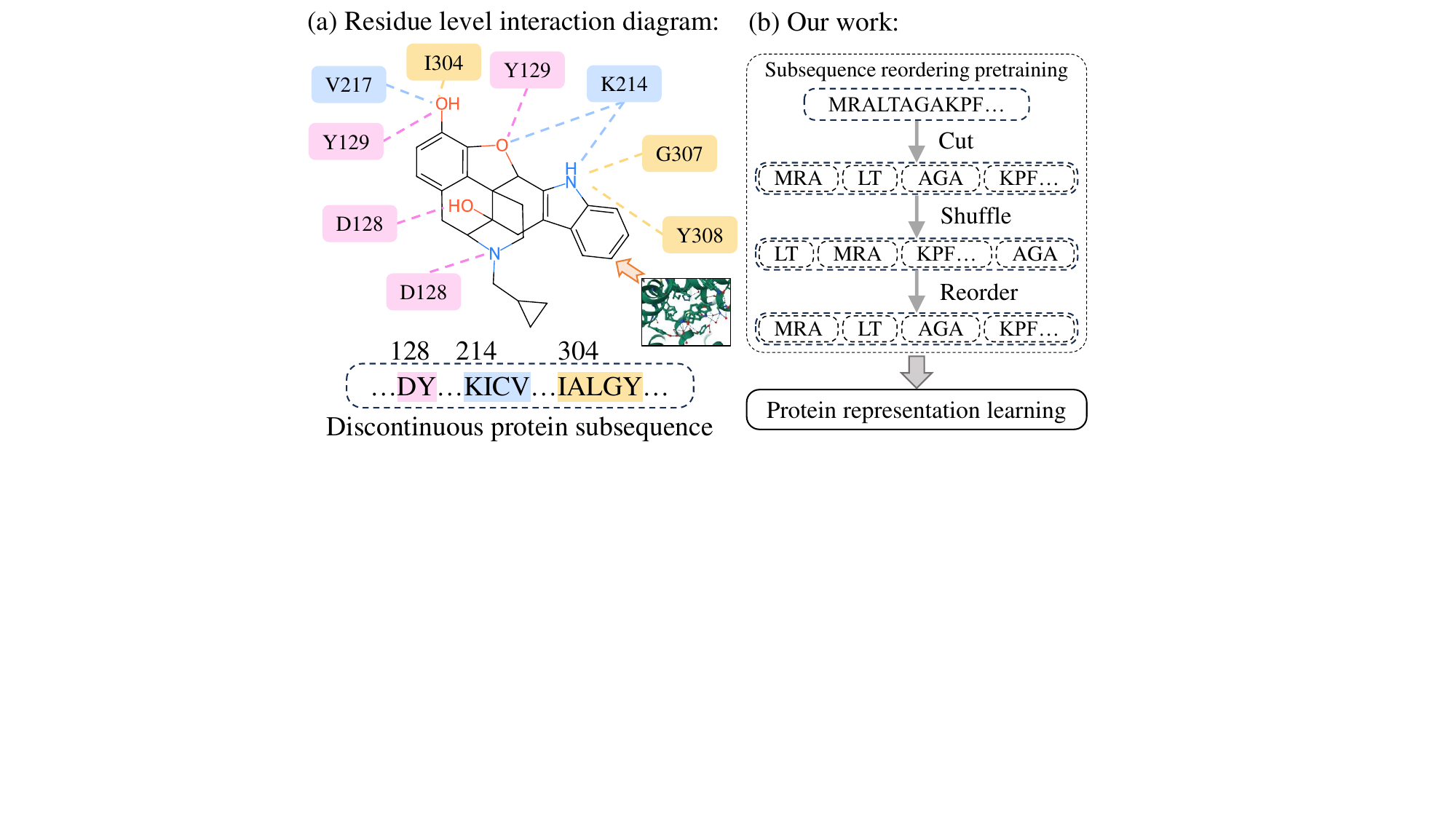}%
  \caption{(a) Residue-level interaction diagram of the EJ4-4N6H protein-ligand complex. (b) Our method learns protein representations by capturing interdependencies among protein subsequences.}
  \label{MotivationFig}
\end{figure}
Compound–protein interaction (CPI) prediction \cite{gao2018interpretable,ding2020identification,gao2024drugclip,wu2024psc} offers a promising solution to the inefficiencies of traditional drug discovery, particularly in reducing time and cost \cite{karimi2020explainable}.
In CPI prediction tasks, evaluation scenarios are typically classified into four categories: Seen-Both, Unseen-Compound, Unseen-Protein, and Unseen-Both based on whether the compounds and proteins involved have appeared in training data. Among these, apart from the Seen-Both scenario, the other three scenarios fall under the category of zero-shot learning. Considering the immense chemical space and the continuous emergence of novel proteins, zero-shot CPI prediction scenarios better reflect the real-world requirements faced in drug discovery.

In the interaction between a compound and a protein, as shown in Fig. \ref{MotivationFig} (a), the amino acid residues involved in binding are often not sequentially adjacent, but are brought into close spatial proximity through the folding of the protein’s three-dimensional structure, thereby forming a binding pocket or active site. 
Therefore, effectively modeling CPIs requires capturing the relationships between protein subsequences that may cooperate in binding. This poses a significant challenge for traditional sequence-based models, which often struggle to model for such long-range and subsequence dependencies.


Supervised learning techniques, including regression \cite{ozturk2018deepdta,nguyen2021graphdta} and classification \cite{lee2019deepconv,tsubaki2019compound} models, have been extensively studied and have demonstrated adequate performance in predicting Seen-Both CPIs, where both compounds and proteins have been observed during training.
However, supervised learning methods that utilize an end-to-end CPI framework for protein representation struggle to effectively capture the intricate interdependencies among protein subsequences, which are crucial for predicting spatial structures and binding properties. Additionally, the predictive performance of these methods heavily relies on a substantial amount of high-quality labeled data. All of the above make it challenging to generalize to zero-shot learning.


Self-supervised learning (SSL) techniques have recently demonstrated excellent performance in zero-shot learning tasks \cite{radford2021learning,robinson2023contrasting,zhao2023gimlet,li2024contrastive}. 
In the context of CPI prediction, SSL-based approaches typically aim to enhance protein representation learning through two dominant paradigms: cross-modal contrastive learning and mask-based sequence reconstruction. 
Despite their success, these methods demonstrate significant limitations in the inadequate modeling of non-local subsequence dependencies.
Contrastive learning methods focus on aligning identical protein subsequences across modalities (e.g., sequence-structure pairs) \cite{wu2024psc}, yet they inherently fail to capture complex interdependencies between distinct subsequences within the same protein. This limitation arises from their reliance on modality alignment objectives, which prioritize global feature consistency over local interaction patterns. Consequently, critical spatial or functional relationships between non-adjacent residues (e.g., discontinuous binding motifs) remain under-represented.
Mask-based methods, particularly protein language models (PLMs) \cite{rives2021biological,ahmed2020prottrans,sledzieski2022adapting}, address local context learning by reconstructing masked amino acids from their neighbors \cite{you2022cross}.  These approaches predominantly model adjacent or proximal subsequences, neglecting long-range dependencies essential for holistic structural and functional representation. 
In addition, SSL-based methods often suffer from a heavy reliance on specialized or task-specific data. For instance, contrastive learning frameworks rely on high-quality multimodal protein data, such as sequence–structure pairs, to capture cross-modal correlations. However, such data are scarce, especially for understudied proteins, limiting their applicability in zero-shot scenarios.
While mask-based methods are less dependent on multimodal data, they still require large-scale pretraining on massive sequence corpora (e.g., UniRef50/90) to learn robust and generalizable protein representations \cite{rives2021biological}.

To address these challenges, we propose a \textbf{P}rotein encoder via  \textbf{S}ubsequence \textbf{R}eordering \textbf{P}retraining method, specifically designed for \textbf{CPI} prediction \textbf{(PSRP-CPI)}. 
First, as shown in Fig. \ref{MotivationFig} (b), PSRP-CPI pretrains the protein encoder via subsequence reordering, explicitly capturing interdependencies among protein subsequences for protein representation learning. Then, it fine-tunes the encoder according to benchmark methods to adapt to the CPI task more effectively.
In the pretraining phase, we use a multi-layer Transformer \cite{vaswani2017attention} as the protein encoder to effectively learn relationships between distant subsequences.
In addition, the reordering task requires the model to predict the correct order of shuffled protein subsequences, forcing it to understand and model their structural and functional interdependencies.
Second, we apply length-variable protein augmentation to improve protein subsequence reordering method's pretraining performance even on small training datasets and enhance the model's zero-shot learning ability. 
Finally, to evaluate the model's validity and zero-shot learning ability, we divided the test data into four groups based on whether compounds and proteins have been observed during training.
The experimental results demonstrate that PSRP-CPI significantly enhances the performance of baseline methods in zero-shot scenarios, highlighting our approach's effectiveness in enhancing model generalization. 
Meanwhile, our approach exhibits enhanced efficacy relative to conventional pre-training methods, especially when operating with constrained training datasets that lack sufficient sample diversity.
The proposed method serves as a powerful tool for CPI tasks, with broad applicability in drug discovery.
The main contributions of this work are as follows:
\begin{itemize}
    \item We propose a pretraining method based on protein subsequence reordering, termed PSRP-CPI, which improves performance on the CPI prediction task by explicitly modeling the relationships between protein subsequences.
    
    \item We develop a length-variable protein augmentation strategy that enables the model to achieve robust pre-training performance, even when trained on a small-scale dataset.
    
    \item We conduct comprehensive experiments on four widely used CPI benchmark datasets to demonstrate the strong performance of PSRP-CPI in zero-shot CPI prediction. Furthermore, we compared PSRP-CPI with existing pretraining methods on small-scale datasets to evaluate its effectiveness in low-resource settings.
\end{itemize}

\section{Related Work}\label{RelatedWork}
\textbf{Compound-Protein Interaction Prediction.}
Clinical and laboratory studies focus on CPI prediction, primarily through molecular docking \cite{lu2022tankbind,pei2024fabind} and representation learning methods \cite{gao2018interpretable,nguyen2021graphdta,chen2020learning}. Molecular docking predicts protein-ligand binding structures, using energy functions and geometric deep learning to rank ligand interactions. However, these methods suffer from low accuracy and high computational costs, limiting their large-scale application.
To address these issues, representation learning methods provide a simpler and more efficient alternative. 
DeepDTA \cite{ozturk2018deepdta} uses convolutional neural networks (CNNs) to extract compound and protein representations, while GraphDTA \cite{nguyen2021graphdta} employs graph neural networks (GNNs) to characterize molecules, marking the first use of GNNs in this domain. HyperattentionDTI \cite{zhao2022hyperattentiondti} and PerceiverCPI \cite{nguyen2023perceiver} introduce attention mechanisms to enable models to perform better in CPI tasks.
In addition, there are methods like DrugVQA \cite{zheng2020predicting}, DrugBAN \cite{bai2023interpretable}, and SiamDTI \cite{zhang2024cross}, use labeled compound-protein pairs to learn the mapping between protein molecular representations by predicting binding probabilities. 
Despite advancements in supervised learning methods employing end-to-end CPI frameworks, current approaches remain limited in their ability to model complex interdependencies among protein subsequences. These structural and functional relationships are critical for accurate prediction of ligand-binding affinities, yet existing architectures often fail to sufficiently encode such subsequence interactions.
This inherent limitation therefore constrains their generalizability when applied to zero-shot prediction scenarios.

\textbf{Pre-training Learning on Proteins.}
Self-supervised learning models are trained by generating supervised information from unlabeled data through pretext tasks, such as text masking, enabling efficient data representation under unsupervised conditions \cite{chen2024self,li2025fspdf}. According to recent studies, self-supervised learning methods, such as context-based and contrastive learning \cite{robinson2023contrasting}, have demonstrated excellent performance in zero-shot scenarios. Models such as ESM \cite{lin2023evolutionary} and ProtBERT \cite{brandes2022proteinbert} employ context-based approaches, masking sequences to train PLMs that predict masked fragments and extract biological protein information. However, these methods focus on the local context, neglecting the interdependencies among the subsequences. They also depend on large-scale unlabeled protein data, requiring substantial data and computational resources.
Contrastive learning has also seen success in compound-protein interaction prediction. For example, DrugClip \cite{gao2024drugclip} uses 3D structural representations of proteins and compounds for comparative learning, offering strong zero-shot generalization. PSC-CPI \cite{wu2024psc} applies in-modal and cross-modality contrastive learning to model protein sequence-structure relationships, improving performance in Unseen-Both generalization. BioCLIP \cite{robinson2023contrasting} performs comparative learning across protein residue and chain levels, boosting downstream task performance. 
However, these methods typically rely on multiple modalities, which are often scarce, limiting their broader applicability. Moreover, contrastive learning methods neglect interactions between distinct subsequences within a protein and fail to represent critical long-range relationships like allosteric communications between distant residues. This limitation may result in challenges when predicting binding sites that involve discontinuous regions.

\begin{figure*}
  \centering
  \includegraphics[width=\linewidth]{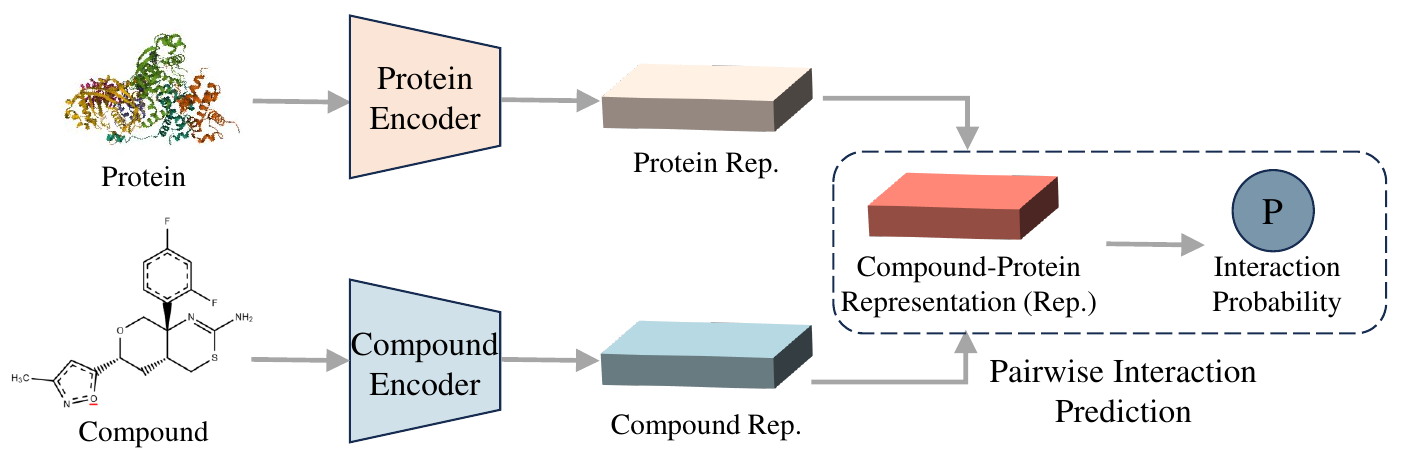}
  \caption{The General Framework for CPI Prediction.}
  \label{pipeline1}
\end{figure*}

\section{Method}\label{Method}
\subsection{Overview}
To state this problem, we will represent the protein of interest as $\mathcal{P}$ and the compound set as $\mathcal{G}$. Given a set of CPI sequences \(E = \{e_{ij} = \{p_i, g_j\} | p_i \in \mathcal{P}, g_j \in \mathcal{G}, Y(e_{ij}) \in \{0, 1\}\}\), where \(Y\) represents a binary indicator. In this scenario, \(Y = 1\) indicates an interaction between the compound and the protein, and \(Y = 0\) implies no interaction. CPI prediction aims to train a model to map the joint feature representation space as an interaction probability score.

Molecules are typically represented using the simplified molecular input line entry system (SMILES). These SMILES representations are subsequently converted into molecular graph structures using tools such as the Deep Graph Library (DGL).
In this case, a compound can be represented by a molecular graph $G_C=(V_C,E_C)$, where each node $a_i \in V_C$ denotes an atom in the compound, and each edge $b_{i,j} \in E_C$ represents a chemical bond between atoms $a_i$ and $a_j$ . Furthermore, a protein with $N_P$ amino acid residues can be represented by a string of its sequence, $S = (r_1, r_2, \cdots , r_{N_P})$, where each residue $r_i$ is one of the 23 amino acid types (all unknown amino acid types are counted as one).
 

\subsection{General CPI Prediction Framework}

Fig. \ref{pipeline1} depicts general CPI prediction framework which consists of three main parts: Encoders that obtain compound and protein representations, and paired interaction prediction modules that predict interaction probabilities.
Next, we introduce compound encoder and paired interaction module, while protein representation learning is discussed in the next subsection.

\textbf{Compound Encoder.}
The encoder takes a molecular graph \( G_C = (V_C, E_C) \) or SMILES string format \( B_C = (A_C) \) as input and learns an \( N \times F \)-dimensional molecular representation, where \( N \) denotes the predefined maximum number of atoms across all molecules. 
In this study, we describe the molecular representations as follows: 
\begin{equation}
Z_{comp} = 
\begin{cases} 
F_c(V_C, E_C), & \text{if input is graph structure} \\
F_c(A_C), & \text{if input is SMILES},
\end{cases}
\end{equation}
where \(F_c \) denotes the compound encoders of the baseline CPI methods, such as graph neural networks \cite{bai2020bi}, convolutional neural networks and Transformer \cite{vaswani2017attention}.

\textbf{Pairwise Interaction Prediction.}
This module utilizes the compound $Z_{comp}$ and protein $Z_{prot}$ representations as input and outputs their fusion representation. The pairwise interaction module is as follows:
\begin{equation}
Z_{joint}=F_{joint}(Z_{comp},Z_{prot}).
\end{equation}
where $F_{joint}$ represents fusion representation learning in baseline methods, such as attentional mechanisms or linear splicing. $Z_{joint}$ represents joint representation.

To calculate the interaction probability, we input the joint representation \( Z_{joint} \) into the decoder, which consists of a fully connected classification layer followed by a sigmoid activation function:
\begin{equation}
p = \mathrm{Sigmoid}(W_fZ_{joint} + b_{joint})\label{con:Sigmoid},
\end{equation}
where \( W_f \) and \( b_{joint} \) represent the learnable weight matrix and bias vector, respectively.

As a downstream task, the primary CPI prediction objective is to estimate the probability of molecular interactions with proteins. Meanwhile, training aims  to minimize the cross entropy loss function:
\begin{equation}
\mathcal{L}=-\sum_i(y_i\log(p_i)+(1-y_i)\log(1-p_i))+\frac\lambda2\|\Theta\|_2^2.
\end{equation}

In this context, \( \Theta \) represents the collection of all learnable weight matrices and bias vectors, \( y_i \) denotes the ground-truth label for the \( i \)-th compound–protein pair, \( p_i \) is the predicted output probability generated by the model, and \( \lambda \) is a hyperparameter controlling the L2 regularization function's strength.


\begin{figure*}[h!]
  \centering
  \includegraphics[width=\linewidth]{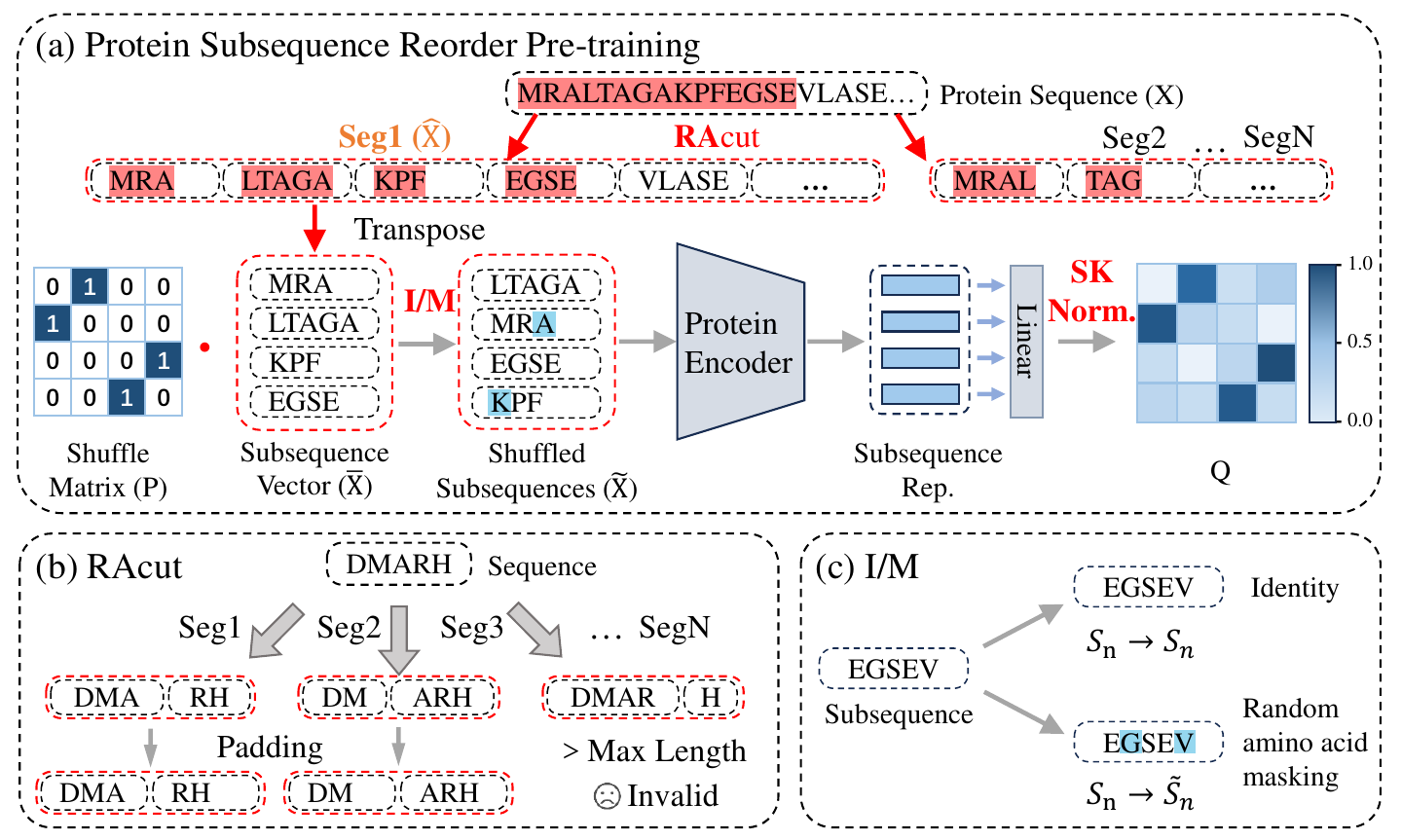}
  \caption{Demonstration of protein subsequence reordering pre-training for compound-protein interaction prediction. (a) First, each protein sequence \(S\) is segmented into multiple subsequences \(\{S_1, S_2, \dots, S_n\}\) using the RAcut strategy. Subsequently, a shuffle matrix \( P \) is applied to randomly rearrange the subsequence order. Then, we use protein encoder and Sinkhorn normalization to transform the subsequence into a doubly-stochastic matrix \( Q \). Finally, the pretraining process is carried out by minimizing the loss between the \( Q \) and \( P \) matrices. (b) The RAcut strategy enables dynamic partitioning of protein sequences into variable-length subsequences, which are then padded to uniform length. (c) Identity transformation (Identity) or random amino acid masking (Mask) operations (I/M) are performed on the subsequences, where the masked amino acids are highlighted in cyan.}
  \label{network}
\end{figure*}

\subsection{Pre-training Methods: Reordering Protein Subsequences}
The purpose of this paper is to propose a method that can obtain the interdependence of protein subsequences.
More importantly, we anticipate that the pre-trained model obtained by this method will effectively address the zero-shot inference challenge.
To achieve this, we propose a protein subsequence reordering pre-training method, as illustrated in Fig. \ref{network}. This method captures the dependencies among subsequences by employing length-variable protein augmentation and subsequence reordering tasks.

\textbf{Length-Variable Protein Augmentation.}
 Data augmentation is crucial for protein representation learning \cite{gao2024drugclip,shen2021improving,sun2024enhancing}. Hence, we introduce a random adaptive cut methodology (RAcut) to maximize extracting the information embedded in the protein sequences, as shown in the Fig. \ref{Method} (b).
Given a protein sequence  \( X \), the RAcut process is as follows: First, the maximum subsequence length \( F_{max} \) is determined based on the maximum allowed length of a protein sequence \( L_{max} \) and number of subsequences \( n \). Next, the protein sequence  \( X \) is segmented into \( n \) parts \(\{S_1, S_2, \dots, S_n\}\) using random lengths, with a minimum subsequence length of 1 and a maximum length of \( F_{max} \). Finally, all subsequences are padded to a uniform length of \( F_{max} \). This cropping strategy is designed to capture protein domains—consecutive subsequences that recur across different proteins and are indicative of their functional roles \cite{ponting2002natural}.

Since RAcut's random subsequence segmentation introduces stochasticity in feature generation, multiple experimental runs may yield minor variations in protein representation. To evaluate the magnitude of this variability, we performed five independent pre-training trials with different random seeds. Experimental results demonstrated that such implementation differences had negligible impact (standard deviation less than 0.3\%) on the performance of downstream CPI tasks, confirming the method's statistical robustness.

 Compared to uniform protein sequence segmentation, this cropping strategy offers two key advantages: First, it enables the model to learn subsequence features with varying lengths, enhancing its ability to perceive proteins across different scales. Second, this method enables multiple segmentations of the same protein sequence, effectively generating diverse training samples from identical sequences. From the perspective of data augmentation, this approach increases the training datasets diversity, this method enhances the diversity of the training dataset and improves the model's zero-shot learning capability.


\textbf{Subsequence shuffle. } The training set \( D = \{(X, P) | X \in \mathcal{T}_c \text{ and } \forall P \in  \mathcal{P}\} \) consists tuples of the original protein sequences \( X \) and shuffle matrices \( P \). Here, the shuffle matrix \( P \in \{0, 1\}^{n \times n} \) is an \( n \times n \) binary matrix in which every row and every column contains exactly one nonzero entry (i.e., a single entry of 1). 
The \( \mathcal{T}_c \) represents a dataset of original protein sequences, which may contain duplicate protein sequences. 
The notation \( \mathcal{P} \) denotes the set of all shuffle matrices. 

Following the RAcut procedure, the original complete protein sequence \( X \) is transformed into a row vector $\hat{X}  = \{S_1, S_2, \dots, S_n\}$, which comprises n subsequences. By transposing $\hat{X}$, we obtain an \(n*1\) column vector $\Bar{X}$, which can then be multiplied with the shuffled matrix \(P\) to yield the shuffled subsequence vector $\tilde{X}$.
Notably, for any given original sequence \( X \), the dataset can be dynamically generated by randomly permuting the order. When combined with the RAcut strategy, this approach yields a substantial amount of data suitable for training large-scale models.

After shuffling the protein subsequences, we introduce a noise function \textbf{I}dentity/ \textbf{M}ask operations (\textbf{I/M}) to generate more diverse subsequences, as shown in the Fig. \ref{network}(c), aligning with common practices in self-supervised learning \cite{chen2020simple,zhangprotein}. This approach enhances the quality of the learned representations, thereby enabling more effective adaptation to zero-shot learning tasks. Specifically, we consider two types of noise functions: \textbf{identity}, which applies no transformations, and \textbf{random amino acid masking}, which randomly masks each amino acid with a fixed probability of 0.15.

\textbf{Protein Encoder.}
We explore the Transformer \cite{vaswani2017attention} model, which has proven to be a highly effective architecture for representation learning tasks, surpassing recurrent and convolutional models in natural language applications. 
The PSRP-CPI employs multilayer Transformer encoder to process protein sequences, using amino acid character sequences as input.
Each input protein sequence is destroyed by randomly shuffling into multiple fragments. 
The Transformer processes the input data by alternating between self-attention mechanisms and feed-forward layers across multiple blocks. The self-attention mechanism enables the model to capture complex representations by integrating contextual information throughout the sequence. 
Then, we employ protein subsequence reordering as the modeling objective to train the model. 
The protein encoder is optimized through sequence reconstruction objectives during pretraining, recovering original sequences from shuffled input sequences. In the inference phase, its parameters remain fixed to assess the validity of representations acquired via the subsequence reordering task.

\textbf{Reordering Protein Subsequences.}
Given a protein sequence ordered by a predetermined standard, we generate a shuffled one by applying a randomly generated shuffle matrix to the original sequence.  Similarly, we can recover the original sequence from a shuffled sequence using the inverse of the respective shuffle matrix. In this case, we define a protein pretraining learning task as one that takes an shuffled sequence as the input and produces a shuffle matrix that scrambles the original sequence as an output.

According to Anfinsen's dogma \cite{Anfinsen}, the tertiary structure of a protein is determined by its amino acid sequence, where the hierarchical assembly of the hydrophobic core strictly depends on the spatiotemporal order of the sequence. By restoring the original arrangement of subsequences, the model implicitly captures global constraints that maintain protein folding feasibility, including: (1) periodic distribution patterns of hydrophobic residues \cite{dill2012protein}, (2) statistical coupling between co-evolving residues \cite{marks2011protein}, and (3) topological compatibility at domain boundaries \cite{moore2008arrangements}.  This process does not directly simulate natural evolutionary pathways but instead learns stable folding rules filtered by evolutionary pressures through data-driven approaches, thereby eliminating invalid combinations that violate either dominant folding thermodynamic trends or functionally essential spatial configurations.

A shuffle matrix \( P \) is a binary square with one non-zero element in each row and column, representing a discrete arrangement of elements.  These matrices play a central role in various formulations but pose challenges for gradient-based optimization methods, which typically rely on continuous updates. Directly manipulating the shuffle matrices is difficult because they are discrete, and optimization algorithms usually require continuous spaces to perform iterative refinements. To address this, we approximately infer the shuffle matrices by considering their nearest convex surrogate, the doubly-stochastic matrix (DSM).

Therefore, we aim to learn a parameterized function \( f_{\theta} : \mathcal{T}_c \ \rightarrow Q \), which maps a fixed-segment protein sequence \( \tilde{X} \) to an \( n \times n \) DSM \( Q \). Ideally, this matrix \( Q \) should correspond to \( P \).
Thus, the shuffle learning problem can be described as:
\begin{equation}
\underset{\theta}{\operatorname*{minimize}}\sum_{(X,P)\in{D}}\Delta\left(P,f_\theta(\tilde{X})\right)+R\left(\theta\right).
\end{equation}
In this expression, \( \tilde{X} \) represents the protein sequence \( X \) rearranged by the shuffle matrix \( P \), \( \Delta(\cdot, \cdot) \) denotes the loss function, \( \theta \) refers to the shuffle learning function's parameters, and \( R(\theta) \) acts as a regularizer to prevent overfitting.

\textbf{Sinkhorn Normalization.}
By solving the sequencing problem, we aim to enable the model capture the protein information behind the sequence. Then, the parametric function $f_{\theta}(\cdot)$ learns intermediate feature representations that encode the amino acid sequence of the input protein data. 
 A matrix $Q~\in~\mathbb{R}_+^{n\times n}$ can be transformed into a doubly stochastic matrix through iterative row and column normalization operations. Therefore, the row \(R(\cdot)\) and column \(C(\cdot)\) normalization operations are defined as follows:
\begin{equation}
R_{i,j}\left(Q\right)=\frac{Q_{i,j}}{\sum_{k=1}^{n}Q_{i,k}};C_{i,j}\left(Q\right)=\frac{Q_{i,j}}{\sum_{k=1}^{n}Q_{k,j}}.
\end{equation}
The Sinkhorn normalization operation \cite{sinkhorn1964relationship,santa2017deeppermnet} for the \(m\)-th iteration can be defined recursively as follows:
\begin{equation}
S^{m}(Q) = \begin{cases} 
Q, & \text{if } m=0, \\
C\left(R\left(S^{m-1}(Q)\right)\right), & \text{otherwise.}
\end{cases}
\label{eq:sinkhorn_iter}
\end{equation}

The Sinkhorn normalization function \( S_{norm} \) is differentiable, and its gradient for the input can be computed efficiently by unrolling the normalization steps. For the row normalization step, the partial derivative can be expressed as follows:

1. Define the normalization denominator for row \( p \):
\begin{equation}
Z_p = \sum_{k=1}^l Q_{p,k}.
\end{equation}

2. The row normalization derivative output \( R_{p,j} \) for \( Q_{p,q} \) is given by:
\begin{equation}
\frac{\partial R_{p,j}}{\partial Q_{p,q}} = \frac{[[j=q]]}{Z_p} - \frac{Q_{p,j}}{Z_p^2},
\end{equation}
where \([[\cdot]]\) is the indicator function, returning \( 1 \) if \( j = q \), and \( 0 \) otherwise.

3. Using the chain rule, the partial derivative of the loss \(\Delta\) for \( Q_{p,q} \) becomes:
\begin{equation}
\frac{\partial \Delta}{\partial Q_{p,q}} = \sum_{j=1}^n \frac{\partial \Delta}{\partial R_{p,j}} \left( \frac{[[j=q]]}{Z_p} - \frac{Q_{p,j}}{Z_p^2} \right).
\end{equation}

The column normalization gradient can be computed analogously by transposing the row and column indices.

As previously stated, our primary objective is to recover the original protein sequence from its shuffled version. 
Intuitively, to recover the precise protein fragment order, the model must identify the dependencies between the individual sequence fragments.




\begin{table}[htbp]
\begin{center}
\caption{Description of CPI datasets.}
\resizebox{0.65\columnwidth}{!}{
\begin{tabular}{lrrr}
\toprule
\textbf{Methods}   & \textbf{Compound} & \textbf{Protein}  & \textbf{Interaction}   \\\midrule
BioSNAP & 4,505 & 2,181 & 27,464 \\
BindingDB  & 14,643 & 2,623 & 49,199  \\
Davis & 68 & 379 & 25,772  \\
KIBA & 225 & 2,068 & 116,350  \\

\bottomrule
\end{tabular}}

\label{tab:datsets}
\end{center}
\end{table}

\section{Experiments}\label{Experiment} 
\subsection{Experimental Setups}

\textbf{Datasets.}
To evaluate the model's effectiveness in the CPI prediction task, experiments are conducted primarily using a publicly available molecular protein dataset, BioSNAP \cite{biosnapnets}, derived from the DrugBank database. This dataset includes 4,510 compounds and 2,181 proteins. It is a balanced dataset, containing positive samples and an equal number of randomly selected negative compound-protein pairs. To better evaluate the model's zero-shot learning ability, we divide the data into four subsets according to the presence of molecules and proteins in the training data, similar to PSC-CPI: (1) Seen-Both (4,166 pairs): both compounds and proteins are seen; (2) Unseen-Comp (5,410 pairs): only proteins are seen; (3) Unseen-Prot (5,082 pairs): only compounds are observed; and (4) Unseen-Both (1,540 pairs): both molecules and proteins are not observed. The other three scenarios, excluding Seenboth, fall under the category of zero-shot learning tasks.
In addition, three common datasets BindingDB \cite{gilson2016bindingdb}, Davis \cite{davis2011comprehensive}, and KIBA \cite{tang2014making} are also used to evaluate the Unseen-Both setting. 
The description about the above four data sets is shown in Tab. \ref{tab:datsets}.
Unlike previous protein pretraining methods that learn transferable knowledge from large amounts of unlabeled data, this paper aims to predict CPI by learning the interdependencies between protein fragments using a small amount of sequence data. Therefore, we perform pretraining solely on the same data provided by the downstream task, without relying on additional unlabeled data.

\textbf{Experimental Settings.} The PSRP-CPI model is trained for 200 epochs with an initial learning rate of 5e-5 and a batch size of 64, using the Adam optimizer. The amino acids utilizes 256-dimensional embeddings, with the protein encoder employing an 8-layer Transformer Encoder. At the fine-tuning stage, the model that achieves the highest area under the receiver operating characteristic curve (AUROC) score on the validation set is selected for evaluation on the test set. The maximum protein sequence length is fixed at 1,200, while the maximum number of atoms in compounds is limited to 290. 
In the protein encoding module, the amino acid type \( N_a \) is set to 23, with all unknown amino acids treated as a single type. We set the number of protein subsequences \(n \) to 24.
The hyperparameters are set to the same for all four datasets.
In addition, to evaluate the PSRP-CPI model's effectiveness, we use the parameter freezing method to fine-tune the CPI task.
Furthermore, the experiments are conducted on a system equipped with an Intel Xeon E5-2690 v3 processor (i.e., 12 cores, 24 threads, and 2.60 GHz) and an RTX 4090 GPU for accelerated computations.
\begin{table*}[!tbp]
\begin{center}
\large
\caption{In the four BioSNAP dataset scenarios, the CPI prediction performance of the existing state-of-the-art (SOTA) methods are evaluated in comparison with its integrated PSRP-CPI module. Positive improvements are highlighted in red, while negative improvements are highlighted in green. }
\resizebox{\textwidth}{!}{
\begin{tabular}{cccccccccc|cc}

\toprule
\multirow{2}{*}{} & \multirow{2}{*}{\textbf{Method}} & \multicolumn{2}{c}{\textbf{Seen-Both}} & \multicolumn{2}{c}{\textbf{Unseen-Comp}} & \multicolumn{2}{c}{\textbf{Unseen-Prot}} & \multicolumn{2}{c}{\textbf{Unseen-Both}} & \multicolumn{2}{c}{\textbf{Avarage}} \\ \cmidrule(r){3-4}  \cmidrule(r){5-6} \cmidrule(r){7-8} \cmidrule(r){9-10} \cmidrule(r){11-12}
 & & \textbf{AUROC} & \textbf{AUPRC} & \textbf{AUROC} & \textbf{AUPRC} & \textbf{AUROC} & \textbf{AUPRC} & \textbf{AUROC} & \textbf{AUPRC} & \textbf{AUROC} & \textbf{AUPRC} \\ \midrule

& DrugBAN & 0.8474 & 0.9066 & 0.7758 & 0.7694 & 0.6916 & 0.5346 & 0.5796 & 0.2610 & 0.7236 & 0.6179 \\
&+PSRP & \textbf{0.8970} & \textbf{0.9297} & \textbf{0.8638} & \textbf{0.8443} & \textbf{0.7446} & \textbf{0.5988} & \textbf{0.6606} & \textbf{0.3559} & \textbf{0.7915} & \textbf{0.6822} \\ 
&$\Delta$ & \textcolor{red}{$5.85\%$} & \textcolor{red}{$2.55\%$} & \textcolor{red}{$11.34\%$} & \textcolor{red}{$9.73\%$} & \textcolor{red}{$7.66\%$} &\textcolor{red}{$12.01\%$} & \textcolor{red}{$13.98\%$} & \textcolor{red}{$36.36\%$} & \textcolor{red}{$9.38\%$} & \textcolor{red}{$10.41\%$} \\ \midrule
& PerceiverCPI & 0.8252 & 0.8974 & 0.8094 & 0.7796 & 0.533 &0.4568 & 0.5216 &0.2850 & 0.6723 & 0.6047 \\
&+PSRP & \textbf{0.8406} & \textbf{0.9010} & \textbf{0.8164} & \textbf{0.7863} & \textbf{0.585} & \textbf{0.4708} & \textbf{0.5570} & \textbf{0.3123} & \textbf{0.6998} & \textbf{0.6176} \\ 
&$\Delta$ & \textcolor{red}{$1.97\%$} & \textcolor{red}{$0.40\%$} & \textcolor{red}{$0.86\%$} & \textcolor{red}{$0.86\%$} & \textcolor{red}{$9.76\%$} &\textcolor{red}{$3.06\%$} & \textcolor{red}{$6.79\%$} & \textcolor{red}{$9.58\%$} & \textcolor{red}{$4.09\%$} & \textcolor{red}{$2.13\%$} \\ \midrule
& SiamDTI & 0.9062 & 0.9438 & 0.8514 & 0.8524 & 0.7393 &0.6172 & 0.6361 & 0.3205 &0.7832 & 0.6835 \\
&+PSRP & 0.9018 & 0.9398 & 0.8431 & 0.8469 & \textbf{0.7765} & \textbf{0.6588} & \textbf{0.6492} & \textbf{0.3473} & \textbf{0.7926} & \textbf{0.6982} \\ 
&$\Delta$ & \textcolor{green}{$0.48\%$} & \textcolor{green}{$0.42\%$} & \textcolor{green}{$0.97\%$} & \textcolor{green}{$0.64\%$} & \textcolor{red}{$5.03\%$} &\textcolor{red}{$6.74\%$} & \textcolor{red}{$2.06\%$} & \textcolor{red}{$8.36\%$} & \textcolor{red}{$1.20\%$} & \textcolor{red}{$2.15\%$} \\ \midrule
& MGNDTI & 0.8987 & 0.9348 & 0.8518 & 0.8354 & 0.7308 & 0.6004 & 0.6008 & 0.2741 & 0.7705 & 0.6612  \\
&+PSRP & \textbf{0.9028} & \textbf{0.9398} &\textbf{0.8519} & \textbf{0.8443} & \textbf{0.7570} & \textbf{0.6169} & \textbf{0.6601} & \textbf{0.3370} & \textbf{0.7929} & \textbf{0.6845}\\  
&$\Delta$ & \textcolor{red}{$0.46\%$} & \textcolor{red}{$0.53\%$} & \textcolor{red}{$0.00\%$} & \textcolor{red}{$1.06\%$} & \textcolor{red}{$3.58\%$} &\textcolor{red}{$2.75\%$} & \textcolor{red}{$9.87\%$} & \textcolor{red}{$22.94\%$} & \textcolor{red}{$2.91\%$} & \textcolor{red}{$3.52\%$} \\ \midrule
& PSC-CPI& 0.8628 & 0.9123 & 0.8018 & 0.7896 & 0.6183 & 0.4155 & 0.5754 & 0.2505 & 0.7146 & 0.5920  \\
&+PSRP & \textbf{0.8679} & \textbf{0.9177} &\textbf{0.8090} & \textbf{0.8062} & \textbf{0.6933} & \textbf{0.5301} & \textbf{0.6171} & \textbf{0.2979} &\textbf{0.7468} & \textbf{0.6380}\\  
&$\Delta$ & \textcolor{red}{$0.71\%$} & \textcolor{red}{$0.59\%$} & \textcolor{red}{$0.90\%$} & \textcolor{red}{$2.10\%$} & \textcolor{red}{$12.13\%$} &\textcolor{red}{$27.58\%$} & \textcolor{red}{$7.25\%$} & \textcolor{red}{$18.92\%$} & \textcolor{red}{$4.51\%$} & \textcolor{red}{$7.77\%$} \\ \bottomrule
\end{tabular}}

\label{tab:FourScenarios}
\end{center}
\end{table*}

\subsection{Comparative Results}
\textbf{Zero-shot Learning Evaluation.}
To evaluate the PSRP-CPI model's effectiveness, we employ the AUROC and area under the precision-recall curve (AUPRC) metrics to evaluate the model's performance in CPI tasks. We conducte five independent rounds of experiments for diverse testing scenarios using distinct random seeds. Furthermore, we report each metric's mean and standard deviation. 
Based on this setup, we integrate PSRP-CPI with other benchmark methods (such as SiamDTI \cite{zhang2024cross} and four other approaches) by replacing the protein encoder with PSRP-CPI to validate its effectiveness.
Our baseline method includes SiamDTI \cite{zhang2024cross}, PerceiverCPI \cite{nguyen2023perceiver}, MGNDTI \cite{peng2024mgndti}, DrugBAN \cite{bai2023interpretable}, and PSC-CPI \cite{wu2024psc}. 

The following observations can be made from Tab. \ref{tab:FourScenarios}:
(1) Effect of pretraining PSRP-CPI: In nearly all scenarios, PSRP-CPI improves the original baseline methods' performance, indicating that our proposed pretraining approach effectively enhances the prediction performance.
(2) Complexity of zero-shot scenarios during CPI tasks: All baseline methods (including those using PSRP-CPI) display varying degrees of performance degradation in different unseen scenarios compared to the Seen-Both setting, highlighting the significant challenges posed by unknown compounds or proteins during CPI tasks.
(3) Zero-shot generalization: The pretrained PSRP-CPI significantly improves the baseline methods' performance in zero-shot scenarios, particularly in the more realistic Unseen-Both scenario.
Under the Unseen-Both scenario, the baseline models achieve a maximum improvement of 13.98\% (with an average improvement of 7.99\%) in AUROC, and a maximum improvement of 36.36\% (with an average improvement of 19.23\%) in AUPRC after integrating the PSRP model.
 These strong performance demonstrates the advantages of our PSRP model.

\begin{table}[!htbp]
\begin{center}

\caption{Under the Unseen-Both setting on two public datasets, the performance of integrating the PSRP module into the current SOTA CPI baseline methods is demonstrated, with improvements highlighted in bold. The results are presented as mean±standard deviation.}
\resizebox{\columnwidth}{!}{
\begin{tabular}{lcccccc}
\toprule
\multirow{2}{*}{\textbf{Methods}} & \multicolumn{2}{c}{\textbf{Bindingdb}} & \multicolumn{2}{c}{\textbf{Davis}}  \\ \cmidrule(r){2-3}  \cmidrule(r){4-5} \cmidrule(r){6-7}
 & \textbf{AUROC} $\uparrow$ & \textbf{AUPRC} $\uparrow$ & \textbf{AUROC} $\uparrow$ & \textbf{AUPRC} $\uparrow$  \\ \midrule

DrugBAN & $0.5268_{\pm0.015}$ & $0.5362_{\pm0.009}$ & $0.6510_{\pm0.027}$ & $0.4150_{\pm0.038}$ \\
+PSRP & $\textbf{0.5592}_{\pm0.014}$ & $\textbf{0.5620}_{\pm0.007}$ & $0.6324_{\pm0.011}$ & $0.3957_{\pm0.016}$ & \\\midrule
PerceiverCPI & $0.5276_{\pm0.004}$ & $0.5364_{\pm0.003}$ & $0.6146_{\pm0.003}$ & $0.4126_{\pm0.014}$ \\
+PSRP & $\textbf{0.5471}_{\pm0.007}$ & $\textbf{0.5475}_{\pm0.009}$ & $\textbf{0.6324}_{\pm0.004}$ & $\textbf{0.4205}_{\pm0.009}$  \\\midrule
SiamDTI & $0.5544_{\pm0.016}$ & $0.5583_{\pm0.015}$ & $0.6548_{\pm0.021}$ & $0.4324_{\pm0.038}$  \\
+PSRP & $\textbf{0.5595}_{\pm0.016}$ & $\textbf{0.5591}_{\pm0.014}$ & $\textbf{0.6840}_{\pm0.026}$ & $\textbf{0.4484}_{\pm0.032}$  \\\midrule
MGNDTI & $0.5238_{\pm0.032}$ & $0.5368_{\pm0.035}$ & $0.6080_{\pm0.039}$ & $0.405_{\pm0.029}$ \\
+PSRP & $\textbf{0.5437}_{\pm0.013}$ & $\textbf{0.5500}_{\pm0.008}$ & $\textbf{0.6446}_{\pm0.030}$ & $\textbf{0.4437}_{\pm0.039}$  \\\midrule
PSC-CPI & $0.5184_{\pm0.018}$ & $0.5305_{\pm0.011}$ & $0.6023_{\pm0.042}$ & $0.3728_{\pm0.025}$ \\
+PSRP & $\textbf{0.5426}_{\pm0.029}$ & $\textbf{0.5433}_{\pm0.023}$ & $0.5965_{\pm0.025}$ & $0.3510_{\pm0.016}$ \\\bottomrule



\end{tabular}} 

\label{tab:UnseenBothComp}
\end{center}
\end{table}

To further evaluate the PSRP-CPI's generalization capability in zero-shot scenarios, we conduct experiments on the Bindingdb and Davis datasets under the Unseen-Both setting, using AUROC and AUPRC as evaluation metrics. 
The experimental results are presented in Table \ref{tab:UnseenBothComp}. According to the table, almost all baseline methods are enhanced when combined with PSRP-CPI. For the Davis dataset, compared to the baseline methods, the performance of DrugBAN and PSC-CPI with integrated PSRP modules slightly decrease. This can be attributed to the smaller scale of the Davis dataset, as the PSRP model's performance fluctuates with small datasets.
For the BindingDB dataset, the performance of all baseline methods integrated with PSRP is enhanced to varying extents.
The aforementioned results demonstrate that the PSCR method exhibits strong generalization capabilities in the zero-shot CPI task by effectively learning the relationships between protein subsequences.

According to Tables \ref{tab:FourScenarios} and \ref{tab:UnseenBothComp}, we observe that the protein representation model pre-trained with multimodal contrast learning such as PSC-CPI \cite{wu2024psc} does not perform adequately during protein structural information loss due to the model overlooking subsequence interactions.

\begin{table*}[!htbp]
\begin{center}
\caption{Comparison experiment results between PSRP-CPI model and other protein language models (PLMs) in the Unseen-Both scenarios of four datasets. The results are the average of five different random seed experiments (measured using AUROC; a higher metric value indicates improved performance). The best result is highlighted in bold.}
\resizebox{\textwidth}{!}{
\begin{tabular}{lccccc}
\toprule
\textbf{Methods}   & \textbf{Pretraining Dataset(Size)} & \textbf{BioSNAP}  & \textbf{Davis}  & \textbf{Bindingdb} & \textbf{KIBA} \\ \midrule
MGNDTI-baseline\cite{peng2024mgndti}  & $-$ & 0.6008 & 0.6080 & $0.5238$ & $0.6377$ \\
PSC-CPI\cite{wu2024psc}  & Karimi-3D(4K) & $0.6498$ & $0.6395$ & $0.5280$ & $0.6534$ \\
ProtST\cite{xu2023protst}  & 540K & $0.6609$ & $0.6523$ & $0.5204$ & $\textbf{0.6630}$ \\
Cross-Interaction\cite{you2022cross} & 59K & $0.6409$ & $0.6664$ & $0.5416$ & $0.6362$ \\
ESM-2\cite{lin2023evolutionary} & UniRef50(24M) & $\textbf{0.6949}$ & $0.6668$ & $0.5038$ & $0.6434$ \\
ProtBert\cite{brandes2022proteinbert} & BFD(2.1B) & $0.6813$ & $\textbf{0.6671}$ & $0.5341$ & $0.6249$ \\
PSRP-CPI(ours)  & 4K & $0.6601$ & $0.6446$ & $\textbf{0.5437}$ & $0.6553$ \\
\midrule
\end{tabular}} 

\label{tab:PreTrainModel}
\end{center}
\end{table*}

\textbf{Comparison with Existing Pretraining Models.}
To evaluate the proposed PSRP method's effectiveness, we compare it with existing pretraining Models, such as ESM-2 \cite{lin2023evolutionary} and ProtBert \cite{brandes2022proteinbert}. These methods are pre-trained on large-scale datasets. During this experiment, the same backbone method is used, and all the pretrained models are fine-tuned with fixed parameters. This approach ensures that the performance during fine-tuning accurately reflects the pretrained model's quality using the Unseen-Both setting. 
As shown in Table \ref{tab:PreTrainModel}, the PLM models display the best performance on the BioSNAP and Davis datasets. This is due to the vast amounts of pretraining data utilized during the experiments.
However, on the larger-scale CPI datasets (i.e., Bindingdb and KIBA) display suboptimal performance and occasional decline. In contrast, the PSRP-CPI method, under the constraints of smaller training datasets, achieves significant improvements on BioSNAP and Davis compared to the backbone model. However, it lags slightly behind the PLM models. Meanwhile, for BindingDB dataset, PSRP-CPI outperforms all other pretraining models, achieving SOTA results. 

Furthermore, we conduct a comparative analysis between our proposed PSRP model and the contrastive learning-based pre-training model PSC-CPI \cite{wu2024psc} and ProtST \cite{xu2023protst}. Limited by the scarcity of protein structure data, PSC-CPI underperforms PSRP across all four datasets. 
ProtST benefits from being trained on large-scale datasets, and its performance is comparable to that of PSRP-CPI.

\begin{table}[htbp]
\begin{center}
\caption{Pre-training methods. We use $f_{\mathrm{residue}}(·)$ to denote the MLP head. CE(·) denotes the cross entropy loss. $h_i^{'}$ denotes the representation of node $i$ after masking the corresponding sampled items in each task. 
$f_i$ denotes the residue type feature.
$h_i^{\text{seq}}$, $h_j^{\text{struct}}$, $h_j^{\text{text}}$ denote the sequence embeddings, structural representations, and biomedical text embeddings of the protein, respectively. $f_{\text{contrast}}$ stands for contrast similarity function. $B$ denotes the index set of samples within a mini-batch.}
\resizebox{0.8\columnwidth}{!}{
\begin{tabular}{lc}
\toprule
\textbf{Method}   & \textbf{Loss function} \\ \midrule
AA-Mask & $\mathcal {L}_i=\mathrm{CE}(f_{\mathrm{residue}}(h_i^{'}),f_i)$  \\
BioTexts Sequences CL  & $\mathcal {L}_i =  -\log \left( \frac{ f_{\text{contrast}}(h_i^{\text{seq}}, h_i^{\text{struct}})}{\sum_{j \in B}  f_{\text{contrast}}(h_i^{\text{seq}}, h_j^{\text{struct}})  } \right)$\\
Sequences Structure CL & $\mathcal{L}_i = -\log \left( \frac{ f_{\text{contrast}}(h_i^{\text{seq}}, h_i^{\text{text}})  }{\sum_{j \in B}  f_{\text{contrast}}(h_i^{\text{seq}}, h_j^{\text{text}}) } \right)$ \\
\bottomrule
\end{tabular}}

\label{tab:LossFunction}
\end{center}
\end{table}
\begin{table*}[!htbp]
\begin{center}
\caption{Comparison experiment results between PSRP-CPI and other pretraining methods in the Unseen-Both scenarios of four datasets. The best result is highlighted in bold.}
\resizebox{\textwidth}{!}{
\begin{tabular}{lcccc}
\toprule
\textbf{Methods}    & \textbf{BioSNAP}  & \textbf{Davis}  & \textbf{Bindingdb} & \textbf{KIBA} \\ \midrule
Baseline   & 0.6008 & 0.6080 & $0.5238$ & $0.6377$ \\
BioTexts Sequences CL  & $0.5539$ & $0.6150$ & $0.5204$ & $0.6317$ \\
Sequences Structure CL & $0.5377$ & $0.6318$ & $0.5255$ & $0.5752$ \\
AA-Mask  & $0.5538$ & $0.6133$ & $0.5095$ & $0.5946$ \\
PSRP-CPI(ours)   & $\textbf{0.6601}$ & $\textbf{0.6446}$ & $\textbf{0.5437}$ & $\textbf{0.6553}$ \\
\midrule
\end{tabular}} 

\label{tab:IdenticalData}
\end{center}
\end{table*}

\textbf{Comparison of limited training data.}
To further validate the effectiveness of our method with limited pretraining data, we compare PSRP with other three pretraining methods under identical data and model conditions. The three pre-training methods include amino acid mask based approach, sequence and structure contrast learning, sequence and biomedical text contrast learning. Their training objectives are summarized in Tab. \ref{tab:LossFunction}. 

We report results in Tab. \ref{tab:IdenticalData}.  Our pretraining method based on subsequence reordering consistently demonstrates superior performance over other methods across all datasets under the same data conditions.
Especially on the BioSNAP dataset, the performance of other pretraining methods is notably inferior to that of benchmark supervised learning methods. This suggests that these pretraining approaches fail to achieve superior representation learning outcomes when data is limited.
On the contrary, our PRSP pre-training method has achieved a nearly 10\% improvement over the benchmark method.
These findings indicate that the proposed PSRP-CPI method exhibits significant advantages for zero-shot tasks, despite utilizing less training data.


\begin{table*}[!htbp]
\begin{center}
\caption{Ablation study on shuffle operation, data augmentation and noisy function used for pre-training(measured using AUROC). The best result is highlighted in bold.}
\resizebox{\textwidth}{!}{
\begin{tabular}{lccccc}
\toprule
\textbf{Methods} & \textbf{Seen-Both}  & \textbf{Unseen-Comp}  & \textbf{Unseen-Prot} & \textbf{Unseen-Both}  & \textbf{Average}\\ \midrule
MGNDTI & $0.8987$ & $0.8518$ & $0.7308$ & $0.6008$  & $0.7705$\\
shuffle+RAcut+I/M(full model)  & $0.9028$ & $0.8519$ & $\textbf{0.7570}$ & $\textbf{0.6601}$ & $\textbf{0.7929}$\\
shuffle+RAcut & $\textbf{0.9039}$ & $\textbf{0.8541}$ & $0.7475$ & $0.6542$  & $0.7899$\\
shuffle & $0.8951$ & $08503$ & $0.7504$ & $0.6467$ & $0.7856$\\
\midrule
\end{tabular}} 

\label{tab:ablation}
\end{center}
\end{table*}
\subsection{Ablation Study}
To evaluate the contribution of different components in our proposed method, we conduct ablation studies on the CPI prediction task. The results are summarized in Tab. \ref{tab:ablation}. Specifically, we compare the MGNDTI model (without PSRP model) against three alternative configurations: (A) the full model, which incorporates pretraining with subsequence shuffle, length-variable augmentation (RAcut), and noisy function (I/M operations); (B) the shuffle + RAcut model, which excludes the I/M operation; and (C) the shuffle-only model, which only includes pretraining with shuffle. Our findings can be summarized as follows: 
(1) Pretraining with shuffle alone significantly improves the performance of the backbone model in the Unseen-Both setting. It is demonstrated that the performance of the model in the zero-shot CPI task can be substantially improved through the extraction of interrelationships among protein subsequences; 
(2) Data augmentation is essential for improving the robustness of CPI prediction models. Our experiments show that RAcut-based augmentation leads to notable performance gains, especially in Seen-Both settings and challenging zero-shot scenarios.
and (3) Introducing noise function further improves the model’s performance in all unknown protein scenarios, which are more consistent with real-world drug development settings. In addition, the combination of RAcut and the noise function enables the generation of more diverse training samples, which facilitates zero-shot learning, as demonstrated in the table.

\begin{figure*}
  \centering
  \includegraphics[width=\linewidth]{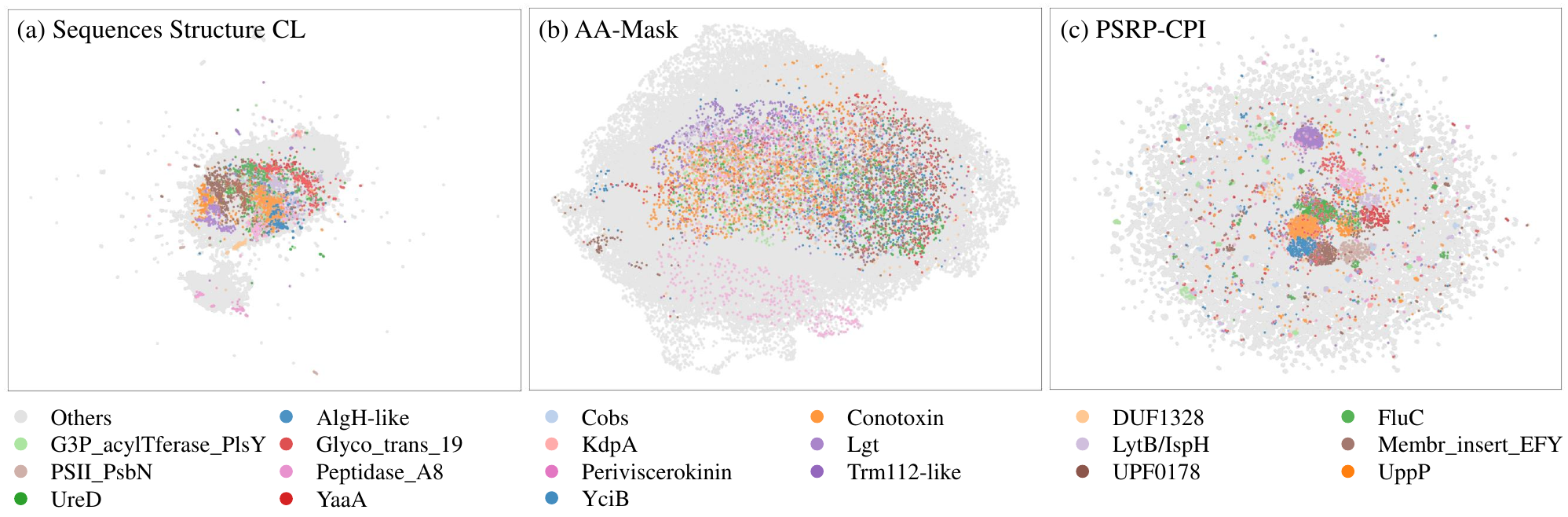}
  \caption{Latent space visualization of protein sequences structure CL (a), mask-base model (b), and PSRP-CPI (c) on UniProt Database.}
  \label{Visualizations}
\end{figure*}
\subsection{Latent Space Visualization}
To qualitatively assess learned protein representation quality, we visualize the latent space of our subsequence reordering-pretrained PSRP-CPI model. Using this model, we extract all protein embeddings from the UniProt Database and project them into 2D space via UMAP \cite{mcinnes2018umap}. Following \cite{akdel2022structural}, the 20 most prevalent protein families are color-coded for visual discrimination. In the comparative analysis, we also visualize the protein representations from the sequences structure contrastive learning-based model and mask-based model pre-trained with the same dataset and model architecture as PSRP-CPI. The visualization results are shown in Fig. \ref{Visualizations}. It is evident that our pre-trained model demonstrates a superior capability in clustering proteins from the same family while effectively separating proteins from different families through greater distances, compared to the other two models. The model's decent capability of discriminating protein families, to some extent, accounts for its superior performance in CPI prediction.

\subsection{Parameter Analysis}
\begin{figure}
  \centering
  \includegraphics[width=\linewidth]{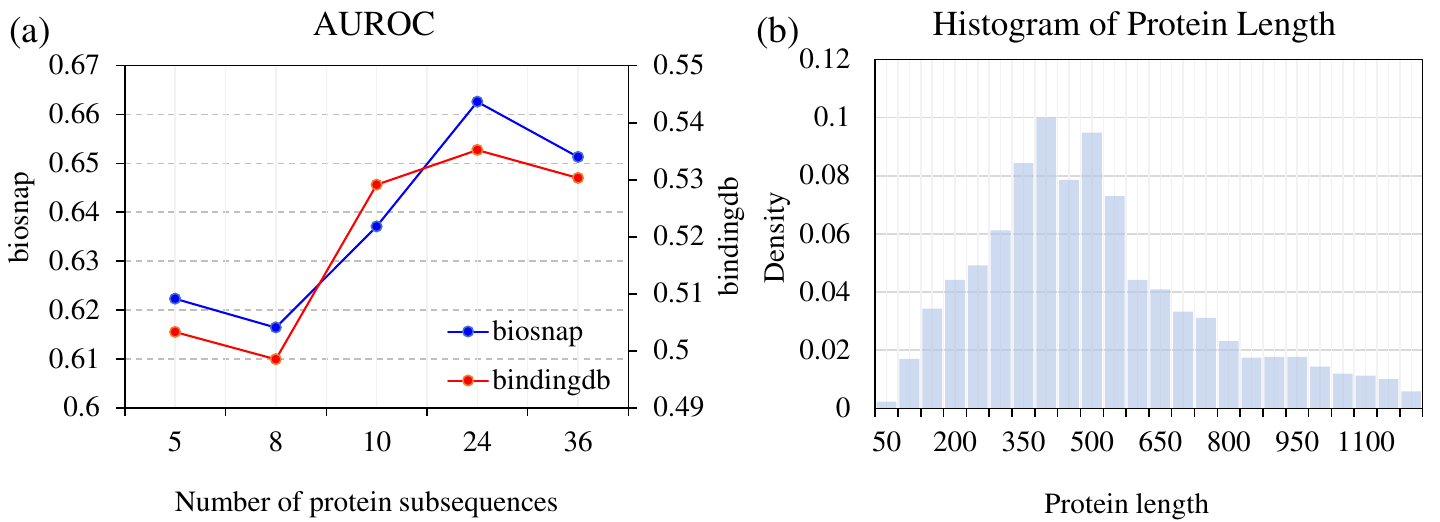}
  \caption{(a) AUROC on two public CPI datasets under the Unseen-Both setting for subsequence cropping. (b) Protein length histogram on the CPI dataset. }
  \label{ParaExp}
\end{figure}

\begin{figure*}
  \centering
  \includegraphics[width=\linewidth]{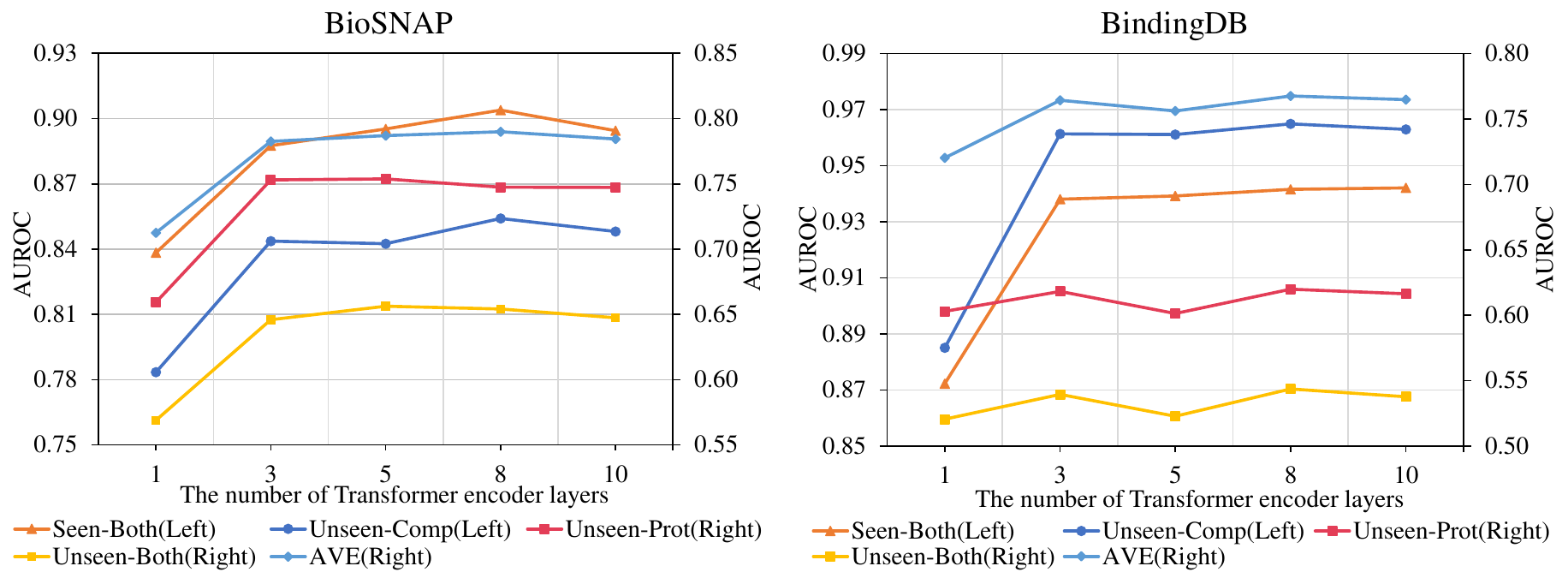}
  \caption{Performance of PSC-CPI about different network layers on BioSNAP and BindingDB.}
  \label{layers}
\end{figure*}

\textbf{Protein Subsequence Quantity: } 
Our random segmentation scheme is designed to extract biologically meaningful, interrelated subsequences for self-supervised learning. To achieve this, determining the number of subsequence segments, \( n \), is crucial. When the number of segments is too high, their lengths become shorter, resulting in insufficient information within and diminishing the inter-segment relationships. This makes it difficult for the model to capture the protein's global features. In contrast, if the subsequence number is too low, the segments may contain excessive redundant information, reducing the model's ability to learn local features effectively.

To investigate the impact of protein segmentation on performance, we employed a subsequence clipping function to demonstrate the experimental results in the Unseen-Both CPI scenario. To mitigate the effects of noisy functions, we only consider identity transformations for the subsequences, as shown in Fig. \ref{ParaExp}(a). For protein sequence segmentation, as the number of segments increases, the performance initially rises and gradually declines, which is consistent with our previous analysis. In practice, we set the number of protein subsequences \(n \) to 24. Based on the sequence length used in training, as shown in Fig. \ref{ParaExp}(b), we hypothesize that an optimal subsequence length of 12 to 20 would enhance pretrained model's performance during the CPI task.

\textbf{Number of Network Layers: }We performe a comprehensive analysis of Transformer architectures with varying depths to assess their effectiveness on prediction tasks. As illustrated in Fig. \ref{layers}, model performance across two datasets generally followed a rise-then-slight-decline pattern as the number of layers increased. The highest average performance across multiple scenarios was observed at a depth of 8 layers. These results indicate that increasing network depth can enhance information propagation and feature extraction, thereby improving predictive performance. However, further deepening the network may introduce excessive representational complexity and over-smoothing effects, ultimately leading to performance degradation.


\section{Conclusion}\label{conclusion}

In this paper, we propose a novel protein pretraining model, PSRP-CPI, for CPI prediction. The PSRP-CPI employs protein subsequence reordering and length-variable augmentation during pretraining. Specifically, PSRP-CPI explicitly models the relationships between protein subsequences, thereby enhancing the model's capacity to perceive binding site information and improving performance with a limited amount of pretraining data. Furthermore, extensive experiments demonstrate that PSRP-CPI significantly improves the performance of existing models on CPI tasks, particularly in the more challenging zero-shot setting. 
Moreover, compared to existing pretraining methods, PSRP-CPI excels with less training data. 

Despite these advancements, there are still limitations. For instance, with the development of protein 3D structure prediction technology, it has become feasible to use 3D structure information to predict CPI. How to incorporate three-dimensional protein structures for pretraining in this task remains an open question. Additionally, due to limited computing resources, we do not use large-scale protein sequence data for pretraining, which presents an opportunity for future work.

\section*{Acknowledge}\label{Acknowledge}
This work was supported in part by the Natural Science Foundation of China (No.62476203), and the Guangdong Provincial Natural Science Foundation General Project (No. 2025A1515012155). Engineering Research Center for Big Data Application in Private Health Medicine of Fujian Universities, Putian University, Putian, Fujian351100, China(MKF202405).











\bibliographystyle{elsarticle-num}   
\bibliography{elsarticle-template-num}   
\end{document}